\documentclass[letterpaper, 10 pt, conference]{ieeeconf}  

\IEEEoverridecommandlockouts                              

\makeatletter
\let\NAT@parse\undefined
\makeatother

\title{\LARGE \bf {Sensor Misalignment-tolerant AUV Navigation with Passive DoA and Doppler Measurements}}
\author{Bingbing Zhang$^{1,2}$, Shuo Liu$^{1,3,5,*}$, Shanmin Zhou$^{4}$, Daxiong Ji$^{1,5}$, Tao Wang$^{1,5,6}$, Tian Xia$^{1}$, Wen Xu$^{1,5}$
	\thanks{This work was supported in part by the 2020 Research Program of Sanya Yazhou Bay Science and Technology City under the grant number SKYC-2020-01-001,  the ``Pioneer'' and ``Leading Goose'' R\&D Program of Zhejiang under the grant number 2023C03124, the Strategic Priority Research Program of the Chinese Academy of Sciences under the grant number XDA22040202, and the Liaoning Natural Science Foundation Program (2022-KF-22-02)}
	\thanks{$^{1}$Key Laboratory of Ocean Observation-Imaging Testbed of Zhejiang Province, Zhejiang University, Zhoushan, 316021, China        {\tt\small \{zhangbb, shuoliu, jidaxiong, twang001, xt0580, wxu\}@zju.edu.cn}}%
	\thanks{$^{2}$Interdisciplinary Student Training Platform for Marine areas, Zhejiang University,
		Hangzhou, 310027, China}%
	\thanks{$^{3}$Hainan Institute, Zhejiang University, Sanya, 572025, China}
	\thanks{$^{4}$Ocean Research Center of Zhoushan, Zhejiang University, Zhoushan, 316021, China {\tt\small timoonboru@foxmail.com}}
	\thanks{$^{5}$The Engineering Research Center of Oceanic Sensing Technology and Equipment, Ministry of Education, Zhoushan, 316000,  China}
	\thanks{$^{6}$State Key Laboratory of Fluid Power and Mechatronic Systems, Zhejiang University, Hangzhou, 310007, China}
	\thanks{*Corresponding author}
}
\usepackage{color}
\usepackage{xcolor}
\usepackage{amsmath,amssymb}
\usepackage{breqn}
\usepackage{mathtools}
\usepackage{layouts}
\usepackage{multicol}
\usepackage{booktabs}
\usepackage{multirow}
\usepackage{url}
\usepackage{threeparttable}
\usepackage[caption=false]{subfig}
\usepackage{siunitx}
\usepackage{bbding}
\usepackage{cases}

\usepackage{algorithm,algpseudocode}

\usepackage{diagbox}
\usepackage[section]{placeins}
\algnewcommand{\And}{\textbf{and}}

\definecolor{red1}{RGB}{165,	96,	39}    
\definecolor{green1}{RGB}{98,125,60}    
\definecolor{blue1}{RGB}{58,89,135}    
\definecolor{pink1}{RGB}{239,175,180}    
\definecolor{yellow1}{RGB}{168,147,40}
\definecolor{black1}{RGB}{85,111,131}

\begin{document}

	\maketitle
	\thispagestyle{empty}
	\pagestyle{empty}

	\begin{abstract}
				We present a sensor misalignment-tolerant AUV navigation method that leverages measurements from an acoustic array and dead reckoned information. Recent studies have demonstrated the potential use of passive acoustic Direction of Arrival (DoA) measurements for AUV navigation without requiring ranging measurements. However, the sensor misalignment between the acoustic array and the attitude sensor was not accounted for. Such misalignment may deteriorate the navigation accuracy. This paper proposes a novel approach that allows simultaneous AUV navigation, beacon localization, and sensor alignment. An Unscented Kalman Filter (UKF) that enables the necessary calculations to be completed at an affordable computational load is developed. A Nonlinear Least Squares (NLS)-based technique is employed to find an initial solution for beacon localization and sensor alignment as early as possible using a short-term window of measurements. Experimental results demonstrate the performance of the proposed method.
	\end{abstract}

\section{INTRODUCTION}
	Autonomous Underwater Vehicles (AUVs) have been the focus of recent developments in marine exploration, and navigation is essential for them to obtain accurate position estimates during deployment, operation, as well as recovery \cite{matsuda2019accurate,fischell2019single}. Nowadays, most AUVs rely on dead reckoning for navigation, but the position estimates are known to drift over time. {}{The state-of-the-art approach still requires absolute acoustic positioning to constrain the error growth of the navigation solution \cite{fischell2019single,choi2022broadband}}. Although the most popular of the acoustic navigation schemes is the Long Baseline (LBL), the cost and time associated with the deployment prevent its use in some of the scientific and civilian applications. {}{The Ultra Short Baseline (USBL), which estimates the position from hybrid integration of angular and ranging measurements \cite{wang2022passive}, is a suitable alternative, for its compact size, easy operation as well as low cost.} 
	
	{}{The use of USBL systems poses some unique challenges. Traditional USBL systems measure two-way travel-time (TWTT) ranges. In these systems, the vehicle must interrogate a beacon in order to obtain a ToF measurement between them, which can be energy-consuming {wang2022passive}. Additionally, in terms of multi-vehicle systems, the communication cost of TWTT increases with the number of AUVs. To address these issues, some of the recent works have developed passive USBL systems that rely on one-way travel-time (OWTT) ranging techniques and  eliminate the need for query pings using synchronized clocks \cite{leonard2016autonomous,7317805}. However, the complexity in the hardware design is increased because both the vehicle and the beacon must be equipped with synchronized stable clock hardware \cite{6504537}. Alternatively, a few studies have demonstrated the potential use of passive acoustic Direction of Arrival (DoA) measurements for AUV navigation without requiring ranging measurements \cite{becker2012simultaneous}. In the bearing, elevation, depth difference (BEDD) relative positioning method \cite{7404453}, the receiver AUV calculates the horizontal range between the sender AUV based on the depth difference and the elevation measurements. The authors of \cite{becker2012simultaneous} and \cite{jia2018simultaneous} have provided different approaches for the problem of simultaneous acoustic beacon localization and AUV navigation with an AUV obtaining DoA-only measurements from a single acoustic beacon, of which the depth is known. However, they assumed that the vehicle is reduced degrees of freedom (DoF), which renders the proposed solutions unsuitable for most underwater vehicles. In \cite{9705991} and \cite{9775432}, a 6-DoF AUV is capable of incorporating the DoA measurements and the dead reckoning positioning for self-localization, which has shown promise, especially in areas of multi-AUV missions. However, the navigation accuracy degrades when the elevation angle is shallow.}
	
	{}{In this paper, we propose a novel algorithm for simultaneous AUV navigation, acoustic beacon localization, and sensor alignment without requiring ranging measurements. Our method is distinct from the existing approaches \cite{becker2012simultaneous,7404453,jia2018simultaneous,9705991,9775432} in that the misalignment of the acoustic array with the attitude sensor is accounted for. Additionally, Doppler speed measurements are incorporated into our system, which improves navigation performance when the elevation angle is shallow and can be used for bridging hardware failures (e.g., DVL outages). }
	
	{}{We highlight our contribution as follows:}
	{}{
	\begin{itemize}
		\item We present a novel algorithm that allows an AUV to perform beacon localization and sensor alignment online, being tolerant to sensor misalignment.
		\item We showcase the performance of the proposed method using experimental results.
	\end{itemize}}
	
	The remainder of the paper is organized in the following way.  Sect. \ref{section3} briefly describes the problem, while Sect. \ref{sectAlgorithm} presents the algorithm. In Sect. \ref{results}, implementation details and experimental evaluations are presented. Finally, the conclusions are given in Sect. \ref{sectConclusion}.

\section{PROBLEM STATEMENT}
\label{section3}
In this section, we present the model of the problem for simultaneous  AUV navigation,  beacon localization, and acoustic array alignment.

The work presented here denotes frame and notation as follows. $\prescript{w}{}{}(\cdot)$ denotes the world coordinate frame, which follows the North, East,
Down (NED) convention. $^{v}(\cdot)$ denotes local vehicle frame, or AHRS frame. $^{a}(\cdot)$ denotes local acoustic array frame. Rotation matrices $ \mathbf{R}$ are used for the convenient rotation of 3-D vectors, while Euler angles can be found in state vectors. {}{{$\mathbf{R}(\boldsymbol{\beta})$} represents the rotation matrix corresponding Euler angle $\boldsymbol{\beta}$.} $(\prescript{w}{a}{\mathbf{R}},   \prescript{w}{}{\mathbf{p}}_{a})$ is acoustic array pose in the world coordinate frame, which transforms points from acoustic array frame to the world coordinate frame. {}{Note that we represent the velocity vector using a bold $\mathbf{v}$, while a non-bold $v$ within scripted notation specifically denotes local vehicle frame. }
\subsection{Measurement}
\label{secMeas}
\begin{figure}[t]
	\centering
	\includegraphics[width=0.7\linewidth]{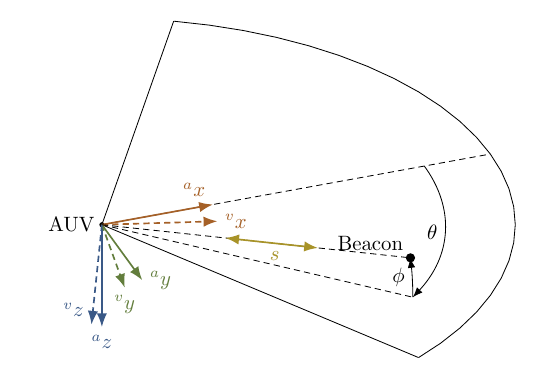}
	\caption{An illustration of the acoustic measurement model. The acoustic array measures the bearing $\theta$ and the elevation $\phi$, as well as the Doppler speed $s$. There is a misalignment between the acoustic frame  $\prescript{a}{}{(\cdot)}$ and the vehicle frame $\prescript{v}{}{(\cdot)}$}.
	\label{model}
\end{figure}

{}{
We assume the vehicle is equipped with an acoustic array, an Attitude and Heading Reference System (AHRS), a DVL, and a pressure sensor. The AHRS measures the attitude, the angular velocity, and the acceleration. The acoustic array is composed of multiple hydrophones and can passively measure the DoA as well as the relative Doppler speed between the vehicle and the beacon. The system within the literature assumes that the AHRS and the DVL are well aligned. There is misalignment between the local acoustic array frame and the AHRS frame, which share the same origin. The beacon is assumed to be equipped with a pressure sensor and fixed in an unknown position. The measurement model of the acoustic array is depicted in Fig. \ref{model}.}
\subsubsection{DoA Measurement}

The beacon location is parameterized as 3D position ($^a\mathbf{p}_b=\left[ ^{a}x_b, \prescript{a}{}{y}_b, \prescript{a}{}{z}_b\right]^\intercal$) in acoustic array frame. 
We assume the DoA measurements at time-step $i$ include two angles (bearing $\theta$ and elevation $\phi$), and can be described by the model: 
\begin{equation}
	\label{eqDoA}
	\mathbf{z}_i =
	\left[\begin{matrix} \theta_i \\ \phi_i \end{matrix}\right] = \left[\begin{matrix} \operatorname{atan2}(^{a}y_{b_i},\prescript{a}{}{x}_{b_i}) \\ \arcsin(\frac{\prescript{a}{}{z}_{b_i}}{||^a\mathbf{p}_{b_i}||}) \end{matrix}\right] + \mathbf{n}_{a_i}
\end{equation}
where $\operatorname{atan2}(y,x)$ \cite{slabaugh1999computing}, calculates arc tangent of the two variables $x$ and $y$, which is similar to calculating the arc tangent of $\frac{y}{x}$ except that the result lies in the range $\left(-\pi,\pi\right]$. {}{$\mathbf{n}_{a_i}$ is the measurement noise vector which is modeled as location-scale t-distribution $\mathcal{LST}(\left[\begin{matrix}
		0\\0\end{matrix} \right] ,\left[\begin{matrix}	\tau_\theta^2& 0\\0  &\tau_\phi^2\end{matrix} \right], 2)$ since the acoustic measurements suffer from underwater inference.}
\subsubsection{Doppler Speed}
The velocity of AUV in the vehicle frame is denoted by $^{v}\mathbf{v}_{i}$.
The acoustic array measures the Doppler speed, which corresponds to the component of $^{v}\mathbf{v}_{i}$ in the direction of the beacon, formulated as
\begin{equation}
	\label{eqDoppler}
	s_i=\frac{{\prescript{v}{}{\mathbf{p}}_{bi}^\intercal }\prescript{v}{}{\mathbf{v}}_i }{||\prescript{v}{}{\mathbf{p}}_{bi}|| } +n_{s_i} 
\end{equation}
where $^v\mathbf{p}_{bi}$ is the beacon location in the vehicle frame and {}{$n_{s_i}$ is assumpted to be location-scale t-distributed noise $n_{s_i}\sim\mathcal{LST}(0, \tau_s^2, 2)$.}
\subsubsection{Depth Measurement}
We assume the beacon is  capable of transmitting the depth information at a  certain rate in real time, so that the beacon depth can be measured by the AUV indirectly as 
\begin{equation}
	d_i=\prescript{w}{}{z}_{bi} +n_{di} 
\end{equation}
where $\prescript{w}{}{z}_b$ is the $z$ component of the beacon position in world coordinate frame $\prescript{w}{}{}\mathbf{p}_{bi}$, $n_{di}$ being Gaussian.
\subsection{Alignment Error}
Since the principal error sources arising in the navigation are alignment angles, we ignore the effect of the translation between the acoustic array and the AHRS. Therefore, a constant rotation matrix $_{a}^{v}\mathbf{R}$, which is computed from three alignment angles (roll $\delta_r$, pitch $\delta_p$ and yaw $\delta_y$), can transform points from acoustic array frame to vehicle frame. $_{a}^{v}\mathbf{R}$ can be calculated by
$
	_{a}^{v}\mathbf{R}=\mathbf{R}(\boldsymbol{\delta})
$, where $\boldsymbol{\delta}$ is the alignment error vector ($\boldsymbol{\delta} = \left[\begin{matrix} \delta_r, \delta_p, \delta_y  \end{matrix}\right]$).

Therefore, the beacon location in the acoustic array frame can be calculated by 
\begin{equation}
	^{a}\mathbf{p}_{bi}={\prescript{v}{a}{\mathbf{R}}^\intercal}   \prescript{v}{}{\mathbf{p}}_{bi}
	\label{eqPb} 
\end{equation}

\subsection{State Estimation}

The problem consists in the real-time estimation of the following state vector at time-step $i$:
\begin{equation}
	\begin{aligned} 
		\boldsymbol{\chi}_i&=
		\left[\begin{matrix} \mathbf{x}_i,	\boldsymbol{\gamma} \end{matrix}\right]^\intercal\\
		\mathbf{x}_i&=\left[\begin{matrix} \prescript{w}{}{}\mathbf{p}_{v_{i}}, \mathbf{r}_{v_{i}} , ^{v}\mathbf{v}_i , ^{v}\mathbf{a}_i , ^{v}\boldsymbol{\omega}_i \end{matrix}\right]^\intercal\\
		\boldsymbol{\gamma}&=\left[\begin{matrix}  \prescript{w}{}{}\mathbf{p}_{b} , \boldsymbol{\delta}  \end{matrix}\right]^\intercal\\
	\end{aligned}
\end{equation}
{}{
where $\mathbf{x}_i$ is the AUV state at time-step $i$} and $\boldsymbol{\gamma}$ contains both  the beacon location  $\prescript{w}{}{}\mathbf{p}_{b}$ and the alignment error vector $\boldsymbol{\delta}$,  which are constant but to be estimated. The AUV state $\mathbf{x}_i$
consists of the position 
$\prescript{w}{}{}\mathbf{p}_{v_{i}}$, the attitude vector $\mathbf{r}_{v_{i}}$ which is composed of three angles (roll, pitch, and yaw), the velocity vector $^{v}\mathbf{v}_i$, the acceleration vector $^{v}\mathbf{a}_i$ and the angular velocity vector $^{v}\boldsymbol{\omega}_i$. 

In order to utilize a state estimator, a state-space representation of the model is required.  Denoting with $\Delta t$ the sampling period of the discrete-time system, the evolution of $\boldsymbol{\chi}_i$ is described as:
\begin{equation}
	\begin{aligned} 
		\prescript{w}{}{}\mathbf{p}_{v_{i+1}}&=\prescript{w}{}{}\mathbf{p}_{v_{i}}
		+ \prescript{w}{v_i}{}\mathbf{R} 
		^{v}\mathbf{v}_i\Delta t+\frac{1}{2} \prescript{w}{v_i}{}\mathbf{R} ^{v}\mathbf{a}_i \Delta t^2
		\\
		\mathbf{r}_{v_{i+1}}&=\mathbf{r}_{v_{i}}+\mathbf{T}(\mathbf{r}_{v_{i}})^{v}\boldsymbol{\omega}_i\Delta t\\ 
		^{v}\mathbf{v}_{i+1}&=^{v}\mathbf{v}_i+^{v} \mathbf{a}_i \Delta t\\
		^{v}\mathbf{a}_{i+1}&=^{v}\mathbf{a}_i\\
		^{v}\boldsymbol{\omega}_{i+1}&=^{v}\boldsymbol{\omega}_i\\
	\end{aligned}
	\label{eqPropagation}
\end{equation}	
where $\prescript{w}{v_{i}}{}\mathbf{R}$ 
represents the rotation matrix corresponding to $\mathbf{r}_{v_{i}}$ and is defined by $\prescript{w}{v_{i}}{}\mathbf{R} =\mathbf{R}(\mathbf{r}_{v_{i}})$. $\mathbf{T}(\mathbf{r}_{v_{i}})$ is given by
\begin{equation}
	\mathbf{T}(\left[r,p,y\right]^\intercal)=\left[\begin{matrix}1& \sin(r)\tan(p)& \cos(r)\tan(p)\\
		0& \cos(r)& -\sin(r)\\
		0& \frac{\sin(r)}{\cos(p)}& \frac{\cos(r)}{\cos(p)}\\
	\end{matrix}\right]
\end{equation}

The available measurements are from the sensors mounted on the vehicle, as is shown in TABLE \ref{tableSensor}.

Since the beacon location in acoustic array frame (Eq. \ref{eqPb}) can be formulated as 
\begin{equation}
	^a\mathbf{p}_{bi}={\prescript{v}{a}{\mathbf{R}}^\intercal } {\prescript{w}{v_{i}}{}\mathbf{R}^\intercal}( \prescript{w}{}{}\mathbf{p}_{b}-\prescript{w}{}{}\mathbf{p}_{v_{i}})
\end{equation}
the DoA measurement (Eq. \ref{eqDoA}) can be associated with some of the states, namely $\prescript{w}{}{}\mathbf{p}_{v_{i}}$,  $\mathbf{r}_{v_{i}}$, $\prescript{w}{}{}\mathbf{p}_{b}$,  as well as $\boldsymbol{\delta}$.  Replacing $\prescript{v}{}{\mathbf{p}}_{b_i}$ with ${\prescript{w}{v_{i}}{}\mathbf{R}^\intercal}( \prescript{w}{}{}\mathbf{p}_{b}-\prescript{w}{}{}\mathbf{p}_{v_{i}})$, the Doppler measurement (Eq. \ref{eqDoppler})  provides observations for the states $\prescript{w}{}{}\mathbf{p}_{v_{i}}$, $\mathbf{r}_{v_{i}}$, $^{v}\mathbf{v}_i$, and $\prescript{w}{}{}\mathbf{p}_{b}$.

%

\begin{table}[htbp]
	\renewcommand\arraystretch{1.2}
	\setlength{\abovecaptionskip}{0pt}%
	\setlength{\belowcaptionskip}{0pt}%
	\caption{THE MEASUREMENTS OF DIFFERENT SENSORS}
	\begin{center}
		\begin{tabular}{p{0.15\linewidth}p{0.17\linewidth}p{0.07\linewidth}p{0.09\linewidth}p{0.25\linewidth}}
			\toprule[1.5pt]
			Sensor&AHRS&DVL&Pressure sensor&Acoustic array\\
			\midrule
			Measurement&$^{v} \mathbf{a}_i$,$^{v}\boldsymbol{\omega}_i$,$\mathbf{r}_{v_{i}}$&$^{v}\mathbf{v}_i$&$\prescript{w}{}{z}_{v_{i}}$&$\mathbf{z}_i$ (Eq. \ref{eqDoA}), $s_i$ (Eq. \ref{eqDoppler})\\
			
			\bottomrule[1.5pt]
		\end{tabular}
	\end{center}
	\begin{tablenotes}
		\footnotesize
		\item[aaa]$\prescript{w}{}{z}_{v_{i}}$ is the $z$ component of the AUV location $\prescript{w}{}{}\mathbf{p}_{v_{i}}$.
	\end{tablenotes}	
	\label{tableSensor}
\end{table}

\section{Algorithm}
\label{sectAlgorithm}
{}{
The proposed approach in this paper is composed of two stages:
\begin{itemize}
	\item{\bf{Step 1}}: Perform state initialization to obtain $\boldsymbol{\gamma}_0$, i.e., the initial values of the beacon location and the alignment error.
	\item{\bf{Step 2}}: Estimate the system states with the basic solution when initialization is completed. 
\end{itemize}}
In this section, we describe the basic solution as well as state initialization in detail. 
\begin{figure}[htbp]
	\centering
	\subfloat[Step 1: Initialization ]{\includegraphics[height=0.55\linewidth]{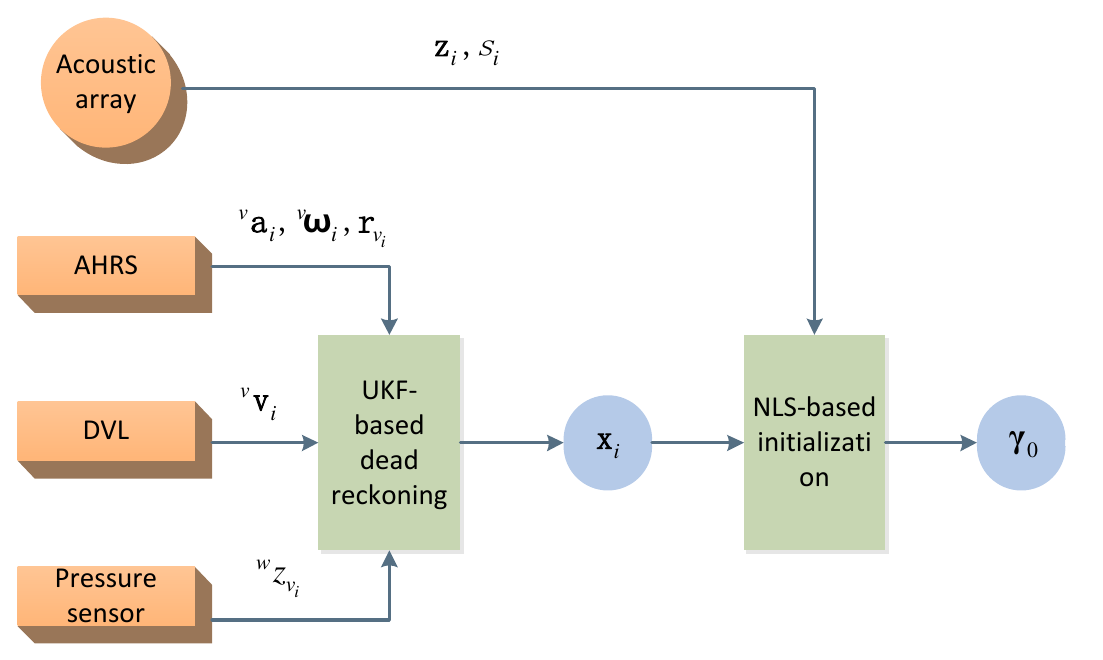}
	}
	\quad 
	\subfloat[Step 2: Basic solution ]{\includegraphics[height=0.5\linewidth]{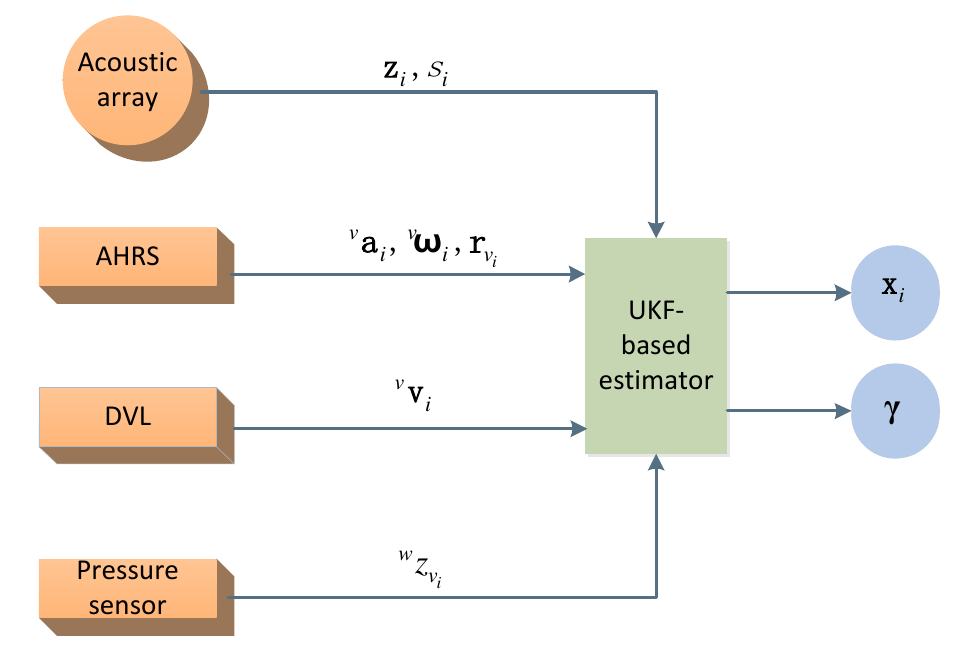}}
	\hfill	
	\caption{Block diagram of the state initialization and the basic solution. Both of the two steps can be performed by AUV in real time}
	\label{figAlgorithm}
\end{figure}
\subsection{Basic solution}

 Considering both the capability of handling non-Gaussian noise 
as well as the nonlinear measurement model, and the computational 
efficiency, the UKF is adopted in this paper, as illustrated in Fig. \ref{figAlgorithm}(b).

\begin{algorithm}
	\caption{Two-step AUV navigation}
	\begin{algorithmic}[1]
		\label{alg:two-stage-auv-nav}
		\State {{}{\textbf{Step 1: State Initialization}}}
		\State Initialize $\mathbf{x}_0$ and $P_{\mathbf{x}_0}$ in UKF
		\State constraints $\gets$ empty list	
		\While{true}
		\State Input DR measurements to UKF
		\State $(\mathbf{x}_i, P_{\mathbf{x}_i}) \gets$ UKF$(\mathbf{x}_{i-1}, P_{\mathbf{x}_{i-1}})$
		\State Add acoustic measurements to constraints
		\State Add $\mathbf{x}_i$ to constraints		
		\If{size(constraints) $>$ desired number}
		\State $\boldsymbol{\gamma}_0 \gets$ NLS(constraints)
		\State \textbf{break}
		\EndIf
		\EndWhile		
		\State Initialize the beacon location and the alignment error $\boldsymbol{\gamma}_0$ and associated covariance $P_{\boldsymbol{\gamma}_0}$ in UKF
		\State {{}{\textbf{Step 2: Basic Solution}}}
		\While{true}
		\State Input DR and acoustic measurements to UKF
		\State $(\boldsymbol{\chi}_i, P_{\boldsymbol{\chi}_i}) \gets$ UKF$(\boldsymbol{\chi}_{i-1}, P_{\boldsymbol{\chi}_{i-1}})$
		\EndWhile
	\end{algorithmic}
	\begin{tablenotes}
		\footnotesize
		\item $\mathbf{x}_i$ is the AUV state, and $P_{\mathbf{x}_i}$ is the associated covariance matrix. Accordingly, $\mathbf{x}_0$ is the initial AUV state, and $P_{\mathbf{x}_0}$ is the initial covariance matrix. DR measurements include AHRS, DVL and pressure sensor measurements, while acoustic measurements include passive DoA and Doppler measurements provided by the acoustic array.
	\end{tablenotes}
	\label{algTwoStep}
\end{algorithm}

\subsection{Initialization}



Suffering from the low update rate of acoustic measurements and the weak observability, the convergence rate of the beacon location and the alignment error is sensitive to the initial state. On the other hand, if we speed the convergence rate by tuning the parameters (e.g.,  initial covariance matrices of the beacon location and the alignment error), the fluctuation of the values is likely to increase, which may in turn affect the accuracy of the navigation system. 

To this end, we propose an NLS-based solution for the initialization of the beacon location and the alignment error, as shown in Fig. \ref{figAlgorithm}(a). Specifically, DoA and Doppler speed measurements as well as AUV states produced by dead reckoning are accumulated over time until sufficient information is available. Then, an NLS-based technique is applied to determine the beacon location and the alignment error.

The problem is formulated as one cost function, which is the sum of the Mahalanobis norm of the prior residual and all measurement residuals: 

\begin{dmath}
	\label{eqNls}
	\min_{\boldsymbol{\gamma}} \quad  \left\{ \textrm{d}_m^2(\boldsymbol{\gamma}-\boldsymbol{\gamma}_0)+\sum_{j\in\mathcal{B}}\textrm{d}_m^2(\mathbf{z}_j-\hat{\mathbf{z}}_j) \\+  \sum_{k\in\mathcal{C}}\textrm{d}_m^2(s_k-\hat{s}_k)  +
	\sum_{l\in\mathcal{D}}\textrm{d}_m^2(d_l-\hat{d}_l)\right\}  \\
\end{dmath}
where $\textrm{d}_m$ calculates the Mahalanobis norm. $\boldsymbol{\gamma_0}$ can be set according to prior knowledge. $\hat{(\cdot)}$ denotes the predicted measurements with the measurement models. The Ceres solver \cite{ceres-solver} is employed to solve this nonlinear problem.

Note that the errors of the AUV states produced by dead reckoning are not considered, because the system with the DVL remains high accuracy in a short period. Also, the beacon location and the alignment error can continuously be corrected with the UKF after initialization. {}{Therefore, the proposed approach in this paper is as described in Algorithm \ref{algTwoStep}. }


\subsection{Observability analysis}
\label{subsec_observability}
{}{System observability examines whether the information from the available measurements is sufficient for estimating the states as well as the parameters without any ambiguity. During the process of state initialization formulated by Eq. \ref{eqNls}, while dead reckoning provides straightforward observations of $\mathbf{x}_i$ (including $ \prescript{w}{}{}\mathbf{p}_{v_{i}}$ and $\mathbf{r}_{v_{i}}$), can the parameters $\boldsymbol{\gamma}$  be identified? The observability analysis of state initialization will be provided as follows.}

\subsubsection{Constraint equation}

With the acoustic DoA measurements in Eq. \ref{eqDoA}, it can be inferred that the beacon lies on a line representing the direction of the signal. According to Eq. \ref{eqPb}, we reformulate the acoustic DoA measurement model in Eq. \ref{eqDoA} as a vector with an uncertain length
\begin{equation}
	\mathbf{m}_i=\lambda_i 	{\prescript{v}{a}{\mathbf{R}}^\intercal } {\prescript{w}{v_{i}}{}\mathbf{R}^\intercal}( \prescript{w}{}{}\mathbf{p}_{b}-\prescript{w}{}{}\mathbf{p}_{v_{i}}),\qquad\lambda_i>0
	\label{eqDoA2}
\end{equation}
where $\lambda_i$ is not available, and two of its perpendicular vectors can be computed by:
\begin{equation}
	\begin{aligned} 
		\mathbf{m}_{i}^{(1)\perp}=\left[\mathbf{m}_i(2),-\mathbf{m}_i(1),0\right]^\intercal\\
		\mathbf{m}_{i}^{(2)\perp}=\left[\mathbf{m}_i(3),0,-\mathbf{m}_i(1)\right]^\intercal
	\end{aligned}
\end{equation}
where $\mathbf{m}_i(k)$ is the k-th element of the vector. Each DoA measurement provides two constraint equations:
${\mathbf{m}_{i}^{(j)\perp}}^\intercal 	{\prescript{v}{a}{\mathbf{R}}^\intercal } {\prescript{w}{v_i}{}\mathbf{R}^\intercal}( \prescript{w}{}{}\mathbf{p}_{b}-\prescript{w}{}{}\mathbf{p}_{v_{i}}) = 0, j = 1,2$.

Since $\boldsymbol{\gamma}$ is six-dimensional, we consider whether three DoA measurements ($\mathbf{M}=\{\mathbf{m}_{i1},\mathbf{m}_{i2},\mathbf{m}_{i3}\}$) can determine $\boldsymbol{\gamma}$ without ambiguity. 
\subsubsection{Analysis based on implicit function theorem}

According to the implicit function theorem, for the given function 
$f(\mathbf{x},\mathbf{y})=\mathbf{0}$
if the Jacobian matrix $\mathbf{Y}$ is an invertible matrix  ($\mathbf{Y}=\frac{\partial f}{\partial \mathbf{y}}$), there is a function $g$ that satisfies $\mathbf{y}=g(\mathbf{x})$ \cite{krantz2002implicit}.

In the problem described above, $f$ is described by 
\begin{subequations}
	\begin{numcases}{}
		{\mathbf{m}_{i1}^{(j)\perp}}^\intercal 	{\prescript{v}{a}{\mathbf{R}}^\intercal } {\prescript{w}{v_{i1}}{}\mathbf{R}^\intercal}( \prescript{w}{}{}\mathbf{p}_{b}-\prescript{w}{}{}\mathbf{p}_{v_{i1}}) = 0, j = 1,2 \notag\\
		{\mathbf{m}_{i2}^{(j)\perp}}^\intercal 	{\prescript{v}{a}{\mathbf{R}}^\intercal } {\prescript{w}{v_{i2}}{}\mathbf{R}^\intercal}( \prescript{w}{}{}\mathbf{p}_{b}-\prescript{w}{}{}\mathbf{p}_{v_{i2}}) = 0, j=1,2\notag\\
		{\mathbf{m}_{i3}^{(j)\perp}}^\intercal 	{\prescript{v}{a}{\mathbf{R}}^\intercal } {\prescript{w}{v_{i3}}{}\mathbf{R}^\intercal}( \prescript{w}{}{}\mathbf{p}_{b}-\prescript{w}{}{}\mathbf{p}_{v_{i3}}) = 0,j=1,2\notag
	\end{numcases} 
\end{subequations}

We define $\boldsymbol{\phi}$ as $
	\boldsymbol{\phi}=\log(\prescript{v}{a}{\mathbf{R}}^\intercal)
$, where $\log$ maps from Lie group $SO(3)$ to  Lie algebra $so(3)$ \cite{barfoot2020state}.

The Jacobian matrix $\mathbf{Y}$ is given by 
\begin{equation}
	\mathbf{Y}_{6\times 6}=\left[\begin{matrix}
		\frac{\partial f}{\partial \boldsymbol{\phi}} &		\frac{\partial f}{\partial \prescript{w}{}{}\mathbf{p}_{b}} 
	\end{matrix}\right]
	\label{eqY}
\end{equation}
where $\frac{\partial f}{\partial \boldsymbol{\phi}}$ can be computed by 
\begin{equation}
	\frac{\partial f}{\partial \boldsymbol{\phi}} = \left[
	\begin{matrix}
		-{\mathbf{m}_{i1}^{(1)\perp}}^\intercal [\prescript{v}{a}{\mathbf{R}}^\intercal {\prescript{w}{v_{i1}}{}\mathbf{R}^\intercal}(\prescript{w}{}{}\mathbf{p}_{b}-\prescript{w}{}{}\mathbf{p}_{v_{i1}})]_{\times}
		\\
		-{\mathbf{m}_{i1}^{(2)\perp}}^\intercal [\prescript{v}{a}{\mathbf{R}}^\intercal {\prescript{w}{v_{i1}}{}\mathbf{R}^\intercal}(\prescript{w}{}{}\mathbf{p}_{b}-\prescript{w}{}{}\mathbf{p}_{v_{i1}})]_{\times}\\
		-{\mathbf{m}_{i2}^{(1)\perp}}^\intercal [\prescript{v}{a}{\mathbf{R}}^\intercal {\prescript{w}{v_{i2}}{}\mathbf{R}^\intercal}(\prescript{w}{}{}\mathbf{p}_{b}-\prescript{w}{}{}\mathbf{p}_{v_{i2}})]_{\times}\\	-{\mathbf{m}_{i2}^{(2)\perp}}^\intercal [\prescript{v}{a}{\mathbf{R}}^\intercal {\prescript{w}{v_{i2}}{}\mathbf{R}^\intercal}(\prescript{w}{}{}\mathbf{p}_{b}-\prescript{w}{}{}\mathbf{p}_{v_{i2}})]_{\times}	\\
		-{\mathbf{m}_{i3}^{(1)\perp}}^\intercal [\prescript{v}{a}{\mathbf{R}}^\intercal {\prescript{w}{v_{i3}}{}\mathbf{R}^\intercal}(\prescript{w}{}{}\mathbf{p}_{b}-\prescript{w}{}{}\mathbf{p}_{v_{i3}})]_{\times}\\	
		-{\mathbf{m}_{i3}^{(2)\perp}}^\intercal [\prescript{v}{a}{\mathbf{R}}^\intercal {\prescript{w}{v_{i3}}{}\mathbf{R}^\intercal}(\prescript{w}{}{}\mathbf{p}_{b}-\prescript{w}{}{}\mathbf{p}_{v_{i3}})]_{\times}
	\end{matrix}
	\right]
\end{equation}
where $\mathbf{v}_{\times}$ is defined by 
\begin{equation}
	\mathbf{v}_{\times}=\left[\begin{matrix}
		0&-\mathbf{v}(3)&\mathbf{v}(2)\\
		\mathbf{v}(3)&0&-\mathbf{v}(1)\\
		-\mathbf{v}(2)&\mathbf{v}(1)&0
	\end{matrix}
	\right]
\end{equation}

In Eq. \ref{eqY}, $\frac{\partial f}{\partial \prescript{w}{}{}\mathbf{p}_{b}}$ is determined by
\begin{equation}
	\frac{\partial f}{\partial \prescript{w}{}{}\mathbf{p}_{b}} = \left[
	\begin{matrix}
		{\mathbf{m}_{i1}^{(1)\perp}}^\intercal \prescript{v}{a}{\mathbf{R}}^\intercal {\prescript{w}{v_{i1}}{}\mathbf{R}^\intercal}\\
		{\mathbf{m}_{i1}^{(2)\perp}}^\intercal \prescript{v}{a}{\mathbf{R}}^\intercal {\prescript{w}{v_{i1}}{}\mathbf{R}^\intercal}\\
		{\mathbf{m}_{i2}^{(1)\perp}}^\intercal \prescript{v}{a}{\mathbf{R}}^\intercal {\prescript{w}{v_{i2}}{}\mathbf{R}^\intercal}\\	{\mathbf{m}_{i2}^{(2)\perp}}^\intercal \prescript{v}{a}{\mathbf{R}}^\intercal {\prescript{w}{v_{i2}}{}\mathbf{R}^\intercal}	\\
		{\mathbf{m}_{i3}^{(1)\perp}}^\intercal \prescript{v}{a}{\mathbf{R}}^\intercal {\prescript{w}{v_{i3}}{}\mathbf{R}^\intercal}\\	{\mathbf{m}_{i3}^{(2)\perp}}^\intercal \prescript{v}{a}{\mathbf{R}}^\intercal {\prescript{w}{v_{i3}}{}\mathbf{R}^\intercal}
	\end{matrix}
	\right]
\end{equation}

With three different DoA measurements, $Rank(\mathbf{Y})=6$ can be achieved, indicating that the parameters $\boldsymbol{\gamma}$ can be observed. However, it must be noted that there exist some exceptions:
\begin{itemize}
	\item If DoA measurements are collected at the same postion $Rank(\mathbf{Y})=5$. 
	\item If three measurements positions and the beacon position are in the same line, satisfying 
	$\prescript{w}{}{}\mathbf{p}_{b}-\prescript{w}{}{}\mathbf{p}_{v_{i1}}\mathop{//}
	\prescript{w}{}{}\mathbf{p}_{b}-\prescript{w}{}{}\mathbf{p}_{v_{i2}}\mathop{//}\prescript{w}{}{}\mathbf{p}_{b}-\prescript{w}{}{}\mathbf{p}_{v_{i3}}$, $Rank(\mathbf{Y})=5$. 
\end{itemize}
That can be explained from a geometric point of view: the measurement $\mathbf{m}_i$ in Eq. \ref{eqDoA2} is related to the direction of $\prescript{w}{}{}\mathbf{p}_{b}-\prescript{w}{}{}\mathbf{p}_{v_{i}}$. In these situations, while this direction can be obtained, there are many solutions of $\prescript{w}{}{}\mathbf{p}_{b}$ satisfying 
$
	\lambda_{i1}(\prescript{w}{}{}\mathbf{p}_{b}-\prescript{w}{}{}\mathbf{p}_{v_{i1}})=\lambda_{i2}(\prescript{w}{}{}\mathbf{p}_{b}-\prescript{w}{}{}\mathbf{p}_{v_{i2}})=\lambda_{i3}(\prescript{w}{}{}\mathbf{p}_{b}-\prescript{w}{}{}\mathbf{p}_{v_{i3}})
$.

\subsubsection{Effects of AUV trajectories}
{}{
So far this section has dealt with whether the parameters can be estimated without ambiguity during state initialization. Another question is what trajectory may enhance observability. }

\begin{figure}[htbp]
	\centering
	\includegraphics[width=0.95\linewidth]{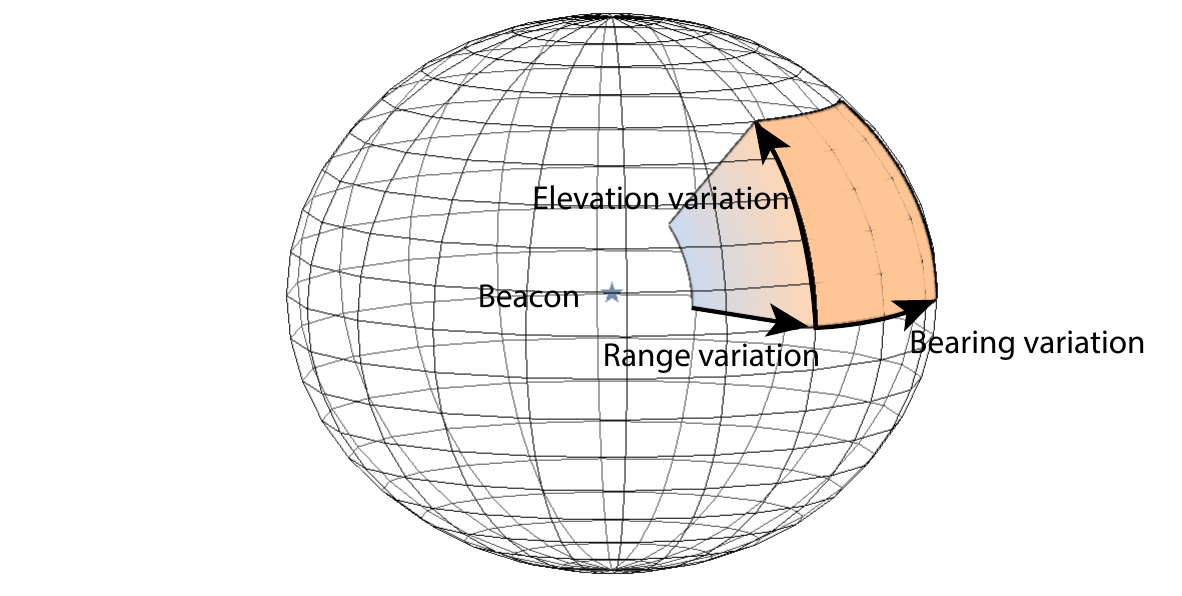}
	\quad 		
	\caption{Distribution of AUV positions during state estimation. We use the spherical coordinate system to code AUV positions with respect to the beacon. Each position is specified by the radial distance $r$ and two angles (i.e., the bearing angle $\prescript{b}{}{}\theta_v$ and the elevation angle $\prescript{b}{}{}\phi_v$ ). Note that the two angles describe the vehicle's direction relative to the beacon.}
	\label{figDemoObs}
\end{figure}

\begin{figure}[htbp]
	\centering
	\includegraphics[width=0.95\linewidth]{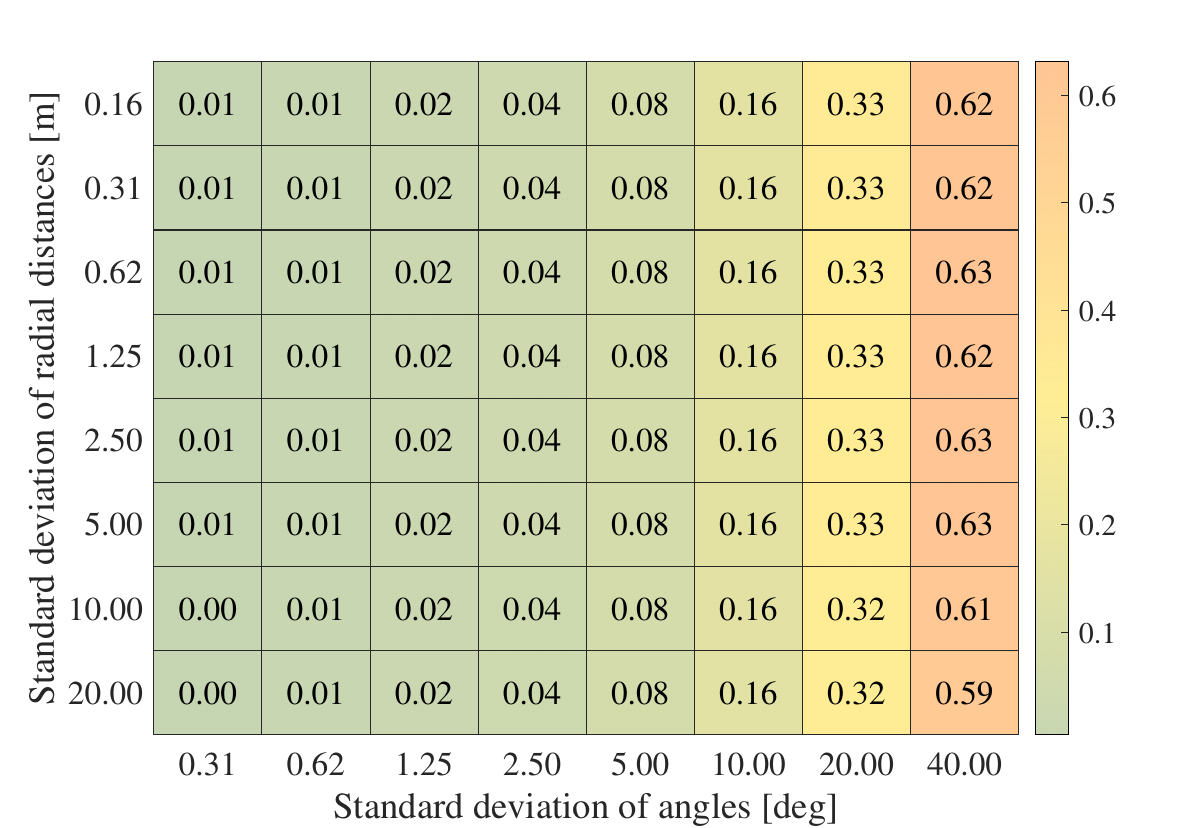}
	\quad 		
	\caption{Simulation results of the effects of range variation and angle variation. We utilize the ratio of the smallest singular value of the Jacobian matrix $\mathbf{Y}$ to the largest singular value to quantitatively evaluate the observability since a smaller ratio indicates that $\mathbf{Y}$ is closer to the situation $Rank(\mathbf{Y})<6$. Here different colors represent different values of this ratio. We denote the standard deviation of angles as $\sigma_a$, and the distribution of two angles follows $\left[\begin{matrix}
			\prescript{b}{}{}\theta_v\\\prescript{b}{}{}\phi_v\end{matrix} \right]\sim \mathcal{N}(\left[\begin{matrix}
			0\\0\end{matrix} \right] ,\left[\begin{matrix}	\sigma_a^2& 0\\0  &\sigma_a^2\end{matrix} \right])$. The distribution of the radial distance obeys $r\sim \mathcal{N}(20, \sigma_a^2)$, where $\sigma_r$ is  the standard deviation of the radial distance.}
	\label{figJacEvaluation}
\end{figure}

{}{Due to the strong nonlinearity of the Jacobian matrix $\mathbf{Y}$ in Eq. \ref {eqY}, analyzing the effects of AUV trajectories on observability is not trivial. To this end, we make some numerical simulations to enrich the understanding of this problem. As illustrated in Fig. \ref{figDemoObs}, we study the effects of different factors on observability. The simulation results are shown in Fig. \ref{figJacEvaluation}, which indicates that the effects of the distribution of the radial distance are negligible. In contrast, the dispersed distribution of angles leads to stronger observability. According to our simulation, the variation of AUV attitudes has little impact on observability, although we do not show the results within the literature.}

\subsection{Outlier rejection}
\label{secOutlier}
{}{
Given the fact that there exists a high probability of outliers in acoustic measurements (i.e., DoA and Doppler measurements), the robustness of state estimation is heavily dependent on outlier rejection. To reduce the impact of outliers, we adopt different strategies in two steps of our algorithm.}

{}{In the first step, i.e., NLS-based state initialization, we utilize Random Sample Consensus (RANSAC) to remove outliers. RANSAC requires multiple iterations to remove outliers, which contributes to its time cost. However, it can provide robustness in the presence of severe outliers in acoustic measurements. In this application, the size of constraints is limited, so RANSAC achieves real-time processing efficiency.}

%

{}{In the second step, i.e., UKF-based full-state estimation, we utilize a naive outlier rejection technique. The measurement is accepted only if the prediction error of the measurement is smaller than three times the square root of the diagonal values of the covariance matrix of the predicted measurement.}


\section{EXPERIMENTAL RESULTS}

\label{results}
In this section, we demonstrate the performance of the proposed method in experiments. {}{We compare it with the following methods:
\begin{itemize} 
	\item Dead reckoning (DR): DR is based on a DVL, an AHRS, and a depth sensor. 
	\item BEDD-aided dead reckoning (abbreviated  as ``BEDD-AID' in the following) \cite{9705991}: BEDD-AID utilizes DoA measurements and depth measurements of the beacon to correct DR. It allows simultaneous AUV navigation and online beacon localization, but does not consider the sensor misalignment. 
	\item USBL-aided dead reckoning (abbreviated  as ``USBL-AID'') \cite{allotta2016new}: USBL-AID utilizes a USBL system to constrain the error growth of position estimates derived from DR. Note that the misalignment of the acoustic array in BEDD-AID has not been discussed in the existing literature, while USBL-AID utilizes an offline method for alignment before operation. We compare the proposed method with USBL-AID in terms of the accuracy of beacon localization and sensor alignment. The latter, which requires the AUV to move along a circle around the beacon on the surface of water, is based on an NLS algorithm and solves the parameters in the following two steps:
	\begin{itemize}
		\item{{Step 1}}: Offline localize the beacon utilizing acoustic ranging and GPS observations:
		\begin{equation}
			\prescript{w}{}{}\mathbf{p}_{b}^* = \mathop{\arg\min}\limits_{\prescript{w}{}{}\mathbf{p}_{b}}\, \sum_{i=1}^N (L_i-||\prescript{w}{}{}\mathbf{p}_{b}-\prescript{w}{}{}\mathbf{p}_{v_{i}}||)^2
		\end{equation}
		where $L_i$ is the range measurement between the AUV and the beacon.  AUV position $\prescript{w}{}{}\mathbf{p}_{v_{i}}$ is provided by GPS.
		\item{{Step 2}}: Offline estimate the misalignment angles based on the estimate of beacon location:
		\begin{equation}
			\boldsymbol{\delta}^* = \mathop{\arg\min}\limits_{\boldsymbol{\delta}}\, \sum_{i=1}^N
			\left\{\textrm{d}_m^2 (\theta_i-\hat{\theta_i}) + \textrm{d}_m^2(\phi_i-\hat{\phi_i}) \right\}
		\end{equation}
		where $\theta_i$ and $\phi_i$ are the angular measurements, while  $\hat{\theta_i}$ and $\hat{\phi_i}$ can be calculated by Eq.\ref{eqDoA} and Eq.\ref{eqPb}, given the position fixes from GPS.
	\end{itemize} 
\end{itemize}}
\subsection{Simulation}

\begin{figure}[htbp]
	\centering
	\subfloat[Trajectory ]{\includegraphics[width=0.95\linewidth]{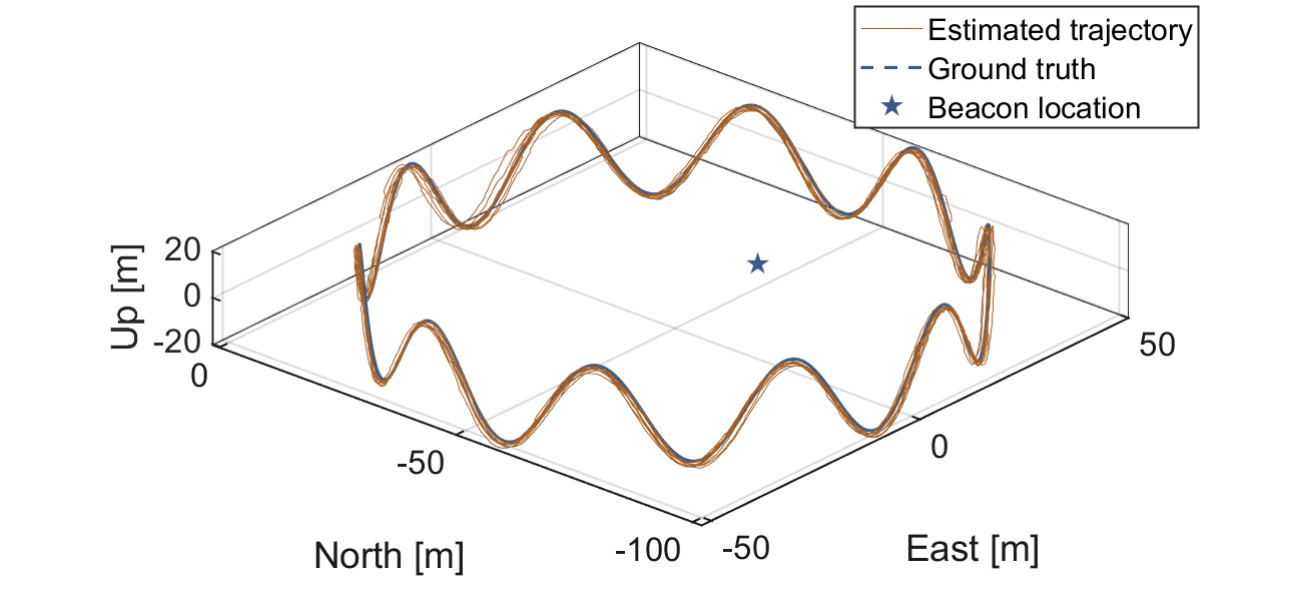}}
	\quad 	
	\subfloat[Attitude ]{\includegraphics[width=0.95\linewidth]{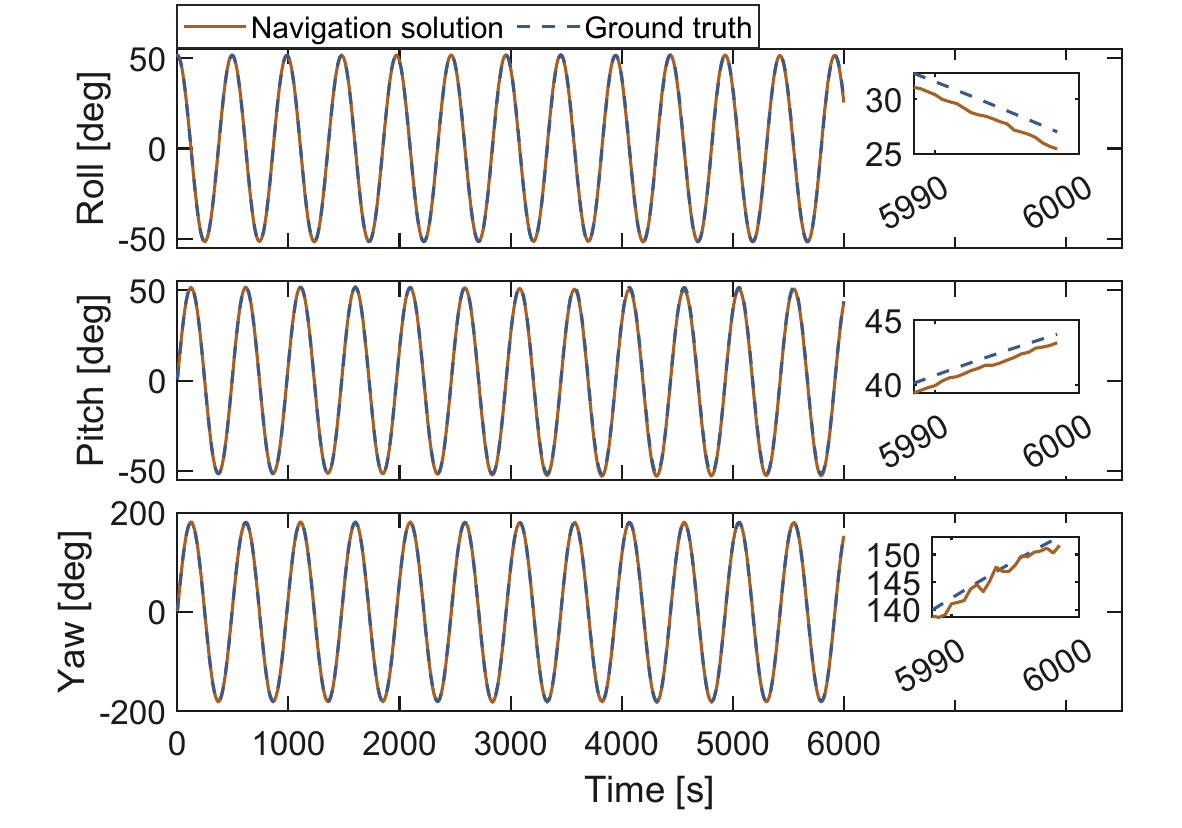}
	}
	\hfill	
	\caption{Trajectory and attitude estimates of the proposed approach versus the ground truth in the simulation. Note that the errors of attitude angles versus the range are minimal, so we zoom in on some parts of the plotted curves.}
	\label{comparision}
\end{figure}

{}{
The simulation setup is outlined in TABLE \ref{SIMULATION}. In the simulation, a  six-degrees-of-freedom AUV moves along a circular trajectory horizontally eight times (the total distance traveled is \SI{7051.41}{\meter}). To thoroughly test the navigation performance of the different methods when operating complex maneuvers, the depth is tasked to vary sinusoidally over time as shown in Fig. \ref{comparision}, and the attitude angles (roll, pitch, and yaw) present significant sinusoidal variation over time. Firstly, we show the accuracy of beacon localization and sensor alignment. Moreover, we demonstrate the overall navigation performance. Additionally, the capability of operating in complex environments with situations of DVL outages is validated. }
\begin{table}[htbp]
	\renewcommand\arraystretch{1.2}
	\setlength{\abovecaptionskip}{0pt}%
	\setlength{\belowcaptionskip}{0pt}%
	\caption{SIMULATION PARAMETERS SETUP}
	\begin{center}
		\begin{tabular}{p{0.15\linewidth}p{0.5\linewidth}c}
			\toprule[1.5pt]
			&Parameter & Value\\
			\midrule
			\multirow{2}{*}{Motion}&Maximum speed of the AUV [m/s]& 1.0\\
			&Duration [$s$] & 6000.0  \\	
			\midrule		
			
			\multirow{3}{*}{\shortstack{AHRS\\ (\SI{20.0}{\hertz})} }&Roll/Pitch measurement noise [$^{\circ}$] & 0.4\\ 
			&Yaw measurement noise [$^{\circ}$] &2.0\\
			&Acceleration noise [m/s$^2$]&0.05\\
			\midrule
			
			\multirow{2}{*}{DVL (\SI{1.0}{\hertz})}& Velocity measurement noise   [m/s]&  0.04 \\
			&Scale error& 1.005\\
			\midrule
			\multirow{3}{*}{\shortstack{Acoustic\\array (\SI{0.2}{\hertz})} }&$\tau_\theta$ and $\tau_\phi$ in DoA measurement  [$^{\circ}$] & 1.0 \\			
			&$\tau_s$ in Doppler speed measurement [m/s] & 0.05   \\		
			\midrule
			\multirow{3}{*}{\shortstack{Constant\\parameters}}& Beacon position ([$\prescript{w}{}{x}_b,\prescript{w}{}{y}_b,\prescript{w}{}{z}_b$]) [m]&[$-50.0,20.0,10.0$]\\
			&Alignment error [$\delta_r,\delta_p,\delta_y$] [$^{\circ} $]&$\alpha$*[$1.0,2.0,3.0$]\\
			\bottomrule[1.5pt]
		\end{tabular}
	\end{center}
	\begin{tablenotes}
		\footnotesize
		\item[aaa]$\tau_\theta$, $\tau_\phi$ and $\tau_s$ are scale parameters of DoA and Doppler measurement noises, which follow location-scale t-distribution as described in Section \ref{secMeas}. The scale error of the DVL is the ratio of the velocity measurement to the actual velocity. $\alpha$ is the misalignment scale.
	\end{tablenotes}	
	\label{SIMULATION}
\end{table}   
\subsubsection{Accuracy of beacon localization and sensor alignment}
We conduct 20 simulated trials for the proposed method and USBL-AID respectively, and the results of beacon localization and sensor alignment are shown in TABLE \ref{beaconlocalization}. While USBL-AID has better accuracy of beacon localization, our method performs better in sensor alignment. That is because the estimation of beacon location is correlated with the navigation error of the AUV, which can be constrained by GPS. USBL-AID utilizes range measurements and GPS position fixes to localize the beacon, decoupling beacon localization and sensor misalignment. On the other hand, our method shows superior performance on sensor alignment compared to that of the USBL-based approach, because the observability is associated with the maneuver of the AUV, and for the latter, the absence of GPS signals underwater prevents the variation of the AUV depth.
\begin{table}[htbp]
	\renewcommand\arraystretch{1.2}
	\setlength{\abovecaptionskip}{0pt}%
	\setlength{\belowcaptionskip}{0pt}%
	\caption{BEACON LOCALIZATION AND SENSOR ALIGNMENT RESULTS}
	\begin{center}
		\begin{tabular}{p{0.14\linewidth}ccc}
			\toprule[1.5pt]
			\multicolumn{2}{c}{Parameter}&Beacon location [m]&Alignment error [deg]\\
			\midrule
			\multirow{2}{*}{USBL-AID}&Mean&[-50.07, 19.69, 10.00]&[2.76, 6.07, 8.93]\\
			&RMSE&0.71&0.75\\
			\cline{2-4}
			\multirow{2}{*}{Proposed}&Mean&[-49.96,19.65,10.00]&[3.00,    5.97,8.97]\\
			&RMSE&0.87&0.13\\			
			\bottomrule[1.5pt]
		\end{tabular}
	\end{center}
	\begin{tablenotes}
		\footnotesize
		\item RMSE is the abbreviation for the root mean square error. Here we set the misalignment scale $\alpha$ in TABLE. \ref{SIMULATION} to $3$, so the ground truth of the alignment error is [3.00, 6.00, 9.00]. 
	\end{tablenotes}	
	\label{beaconlocalization}
\end{table}
	

	\subsubsection{Navigation performance}
\begin{figure}[htbp]
		\centering
		\includegraphics[width=0.95\linewidth]{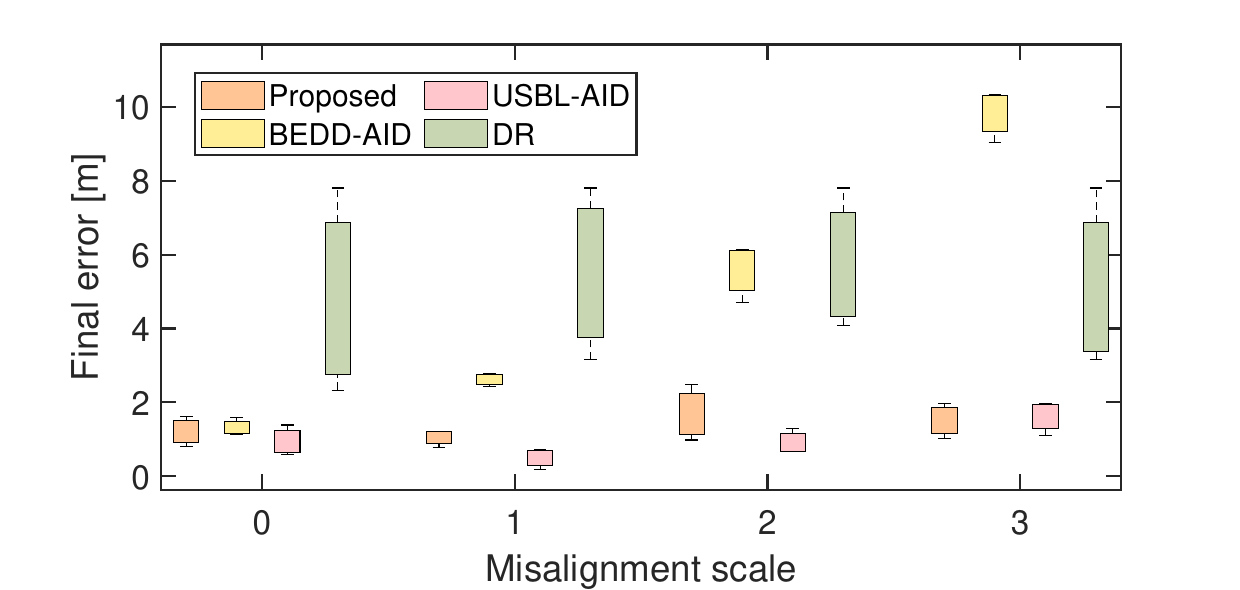}
		\caption{The navigation error of different methods with respect to different extents of sensor misalignment. We change the magnitude of misalignment by setting different misalignment scale values in TABLE \ref{SIMULATION}.}
		\label{boxplot}
	\end{figure}

	{}{
	The navigation performance of our approach is illustrated in  Fig. \ref{comparision}. We compare the navigation error of different methods in Fig. \ref{boxplot}.  Obviously, the proposed method shows the ability to achieve constrained error growth in position estimates compared to DR. When the sensor misalignment is minimal, BEDD-AID leads to better navigation performance, which is close to the proposed method. Larger misalignment can deteriorate the navigation accuracy of BEDD-AID, but has negligible impact on the proposed method. }
	
	\subsubsection{Capability of bridging DVL outages}
\begin{figure}[htbp]
		\centering
		\includegraphics[width=0.95\linewidth]{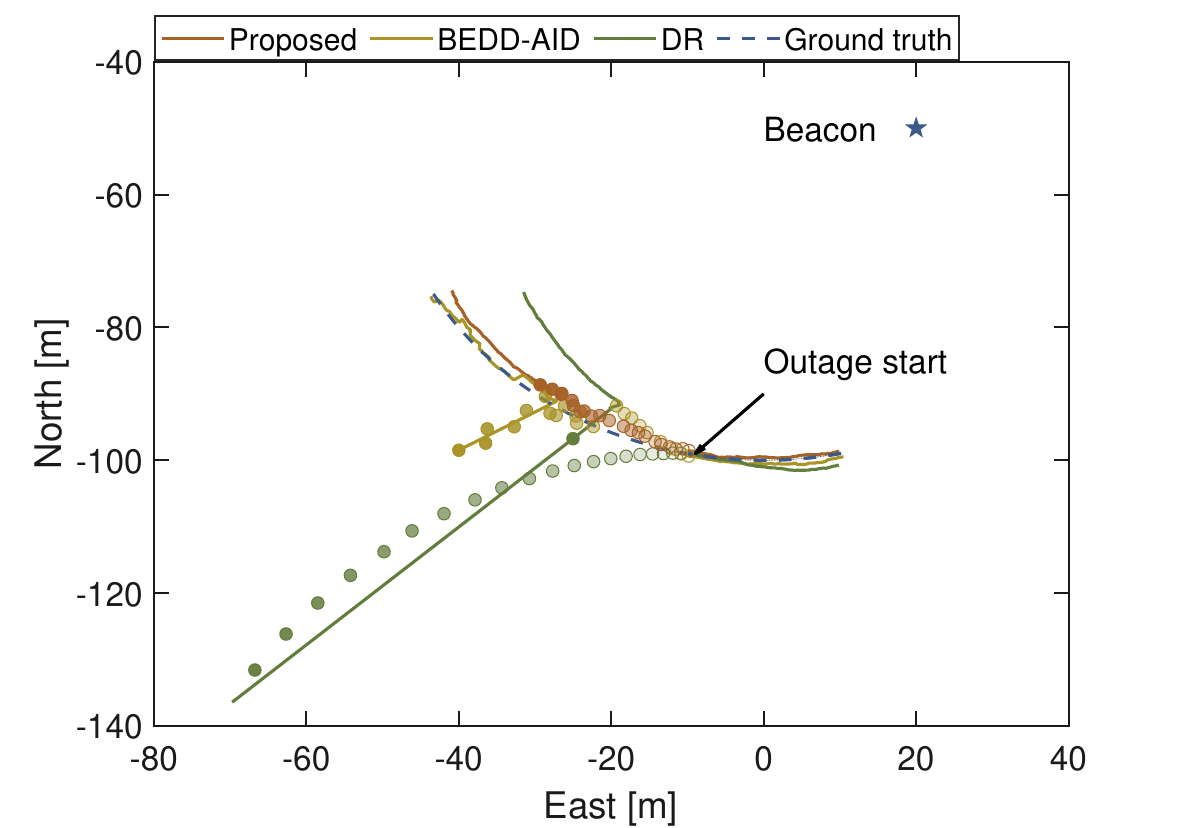}
		\quad 	
		
		\caption{Comparison between the performance of bridging DVL outages with different methods. DVL outages are from \SI{1150}{\second} to \SI{1200}{\second}. The solid lines represent the navigation solutions where DVL is available, while the circles are the results when DVL outages happen. The transparency of the circles,  changing from  $\circ$ to $\bullet$, indicates the evolution of the navigation solutions. }
		\label{figTrajOutage}
	\end{figure}

	{}{
	When no DVL measurements are available, DR will rely only on the AHRS. As a result, the velocity errors as well as the localization errors will rapidly diverge. We simulate DVL outages from \SI{1150}{\second} to \SI{1200}{\second}, and the state estimates from \SI{1100}{\second} to \SI{1250}{\second} were sampled. The results of the proposed method, DR, and BEDD are presented in Fig. \ref{figTrajOutage}. It can be seen that the proposed method is much more robust to  DVL outages than BEDD-AID. That is because in this simulation, the elevation angle $\prescript{b}{}{}\phi_v$ is shallow. This results in the horizontal range estimates being prone to random and systematic errors in BEDD-AID \cite{9775432},  since the horizontal range is observed by multiplying the  depth difference with $\cot(\prescript{b}{}{}\phi_v)$ and a minimal error in $\prescript{b}{}{}\phi_v$ leads to larger errors when it is shallow. On the other hand, the Doppler measurements in the proposed method, provide the observation of the variation in the radial distance $r$. The radial distance measures the horizontal distance by $\sin(\prescript{b}{}{}\phi_v)$, which is robust to the magnitude of $\prescript{b}{}{}\phi_v$. Therefore, the Doppler measurements improve the estimation accuracy of the horizontal range.	}

	\begin{figure}[htbp]
		\centering
\includegraphics[width=0.95\linewidth]{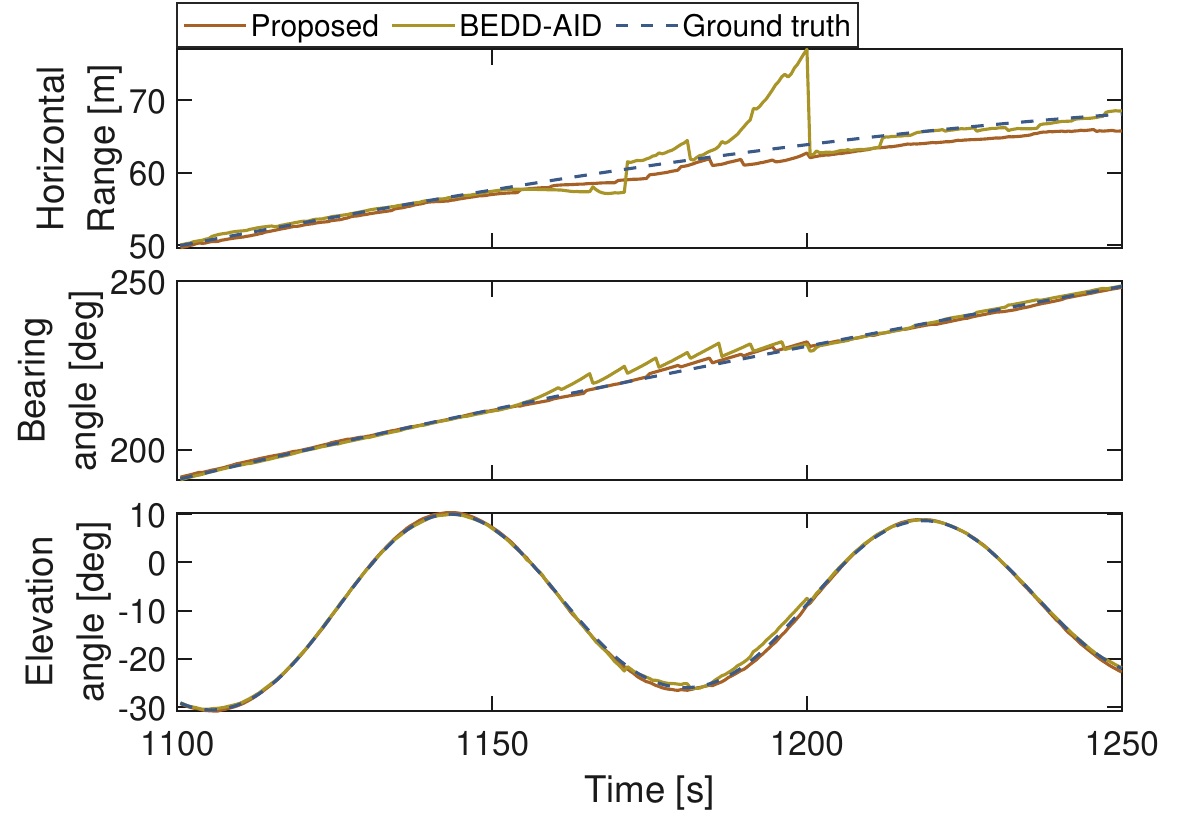}
			\label{outageb}
		\caption{Comparison between the performance of the proposed method and BEDD-AID  in situations where DVL is not available. The representation of the bearing angle $\prescript{b}{}{}\theta_v$ and the elevation angle $\prescript{b}{}{}\phi_v$ are illustrated in Fig. \ref{figDemoObs}.}
		\label{figDoaRange}
	\end{figure}	


		\addtolength{\textheight}{-0.0cm}   

\subsection{Preliminary Field Test}
\begin{figure}[htbp]
	\centering
	\subfloat[Haihong \uppercase\expandafter{\romannumeral1} payload]{\includegraphics[width=0.49\linewidth]{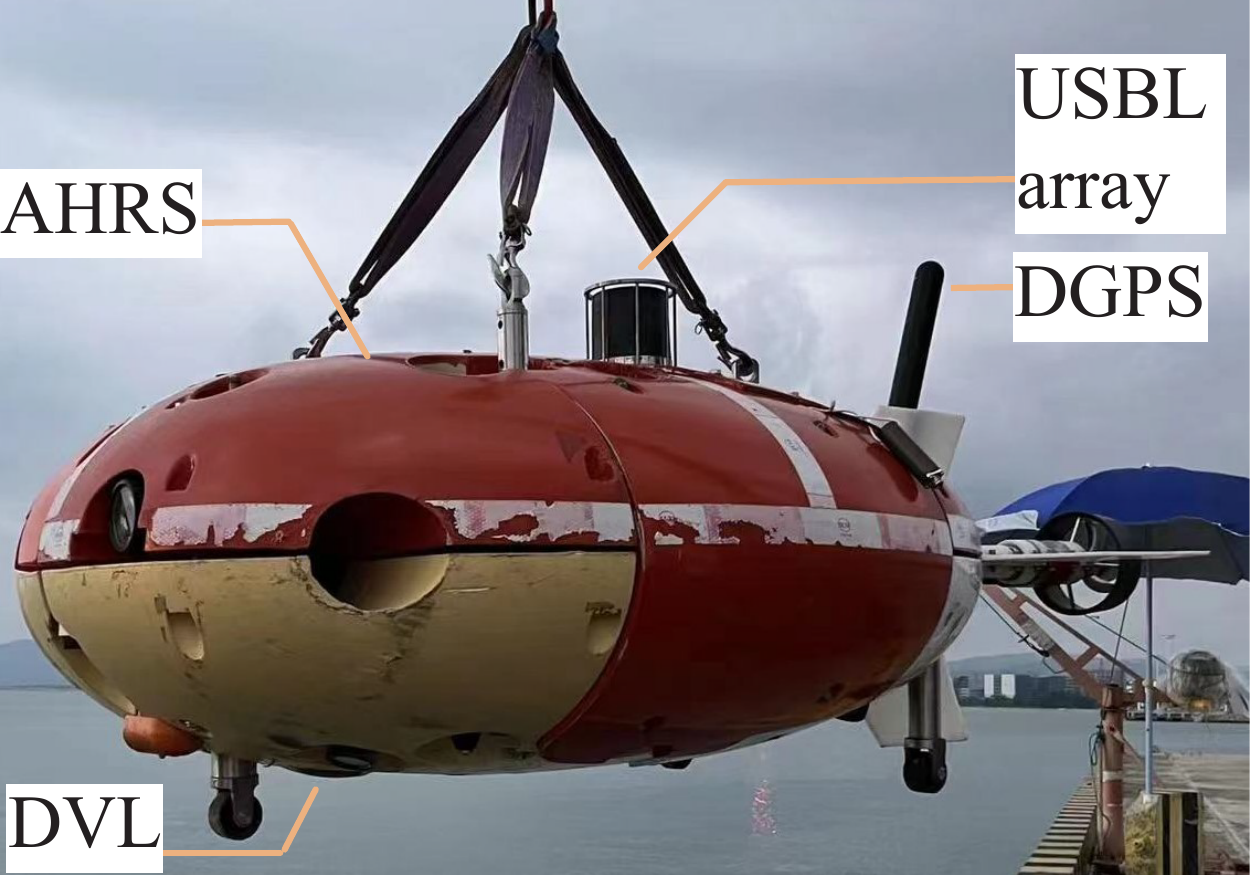}}
	\subfloat[Beacon moored near the surface ]{\includegraphics[width=0.49\linewidth]{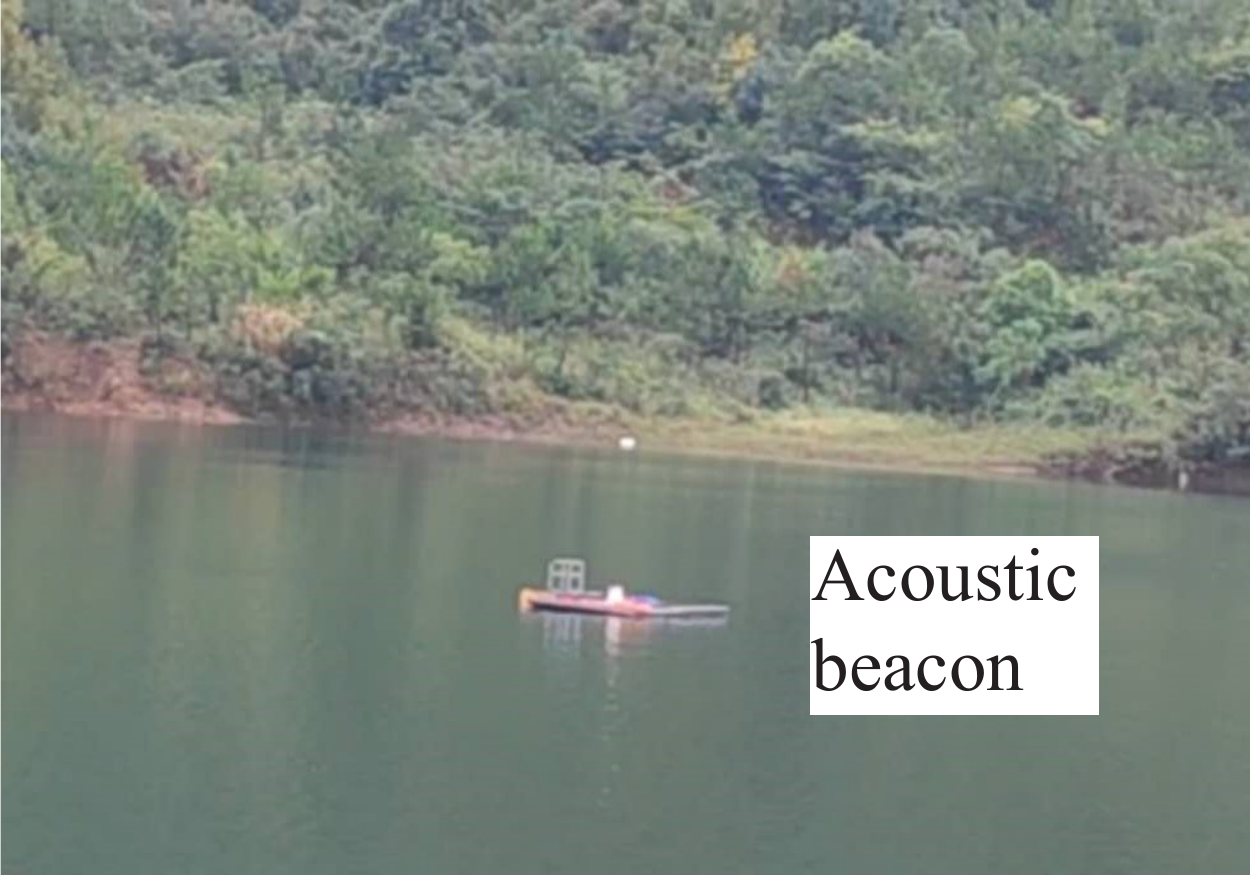}}
	\label{outageb}
	\caption{Experimental setup in Maishan reservoir, Zhoushan, China}
	\label{auv}
\end{figure}

		\addtolength{\textheight}{-0.0cm}   
{}{
We conduct preliminary field tests using Haihong  \uppercase\expandafter{\romannumeral1} AUV (shown in Fig. \ref{auv}) in Maishan reservoir, Zhoushan, China. Different methods have been tested using a collected dataset. As depicted in Fig. \ref{auv}, the AUV is equipped with Xsens, MTi-630 AHRS, Rowe Technologies, Inc., SeaPilot DVL \cite{dvl}, EvoLogics, S2CR USBL \cite{usbl}, as well as Differential Global Positioning System (DGPS) module. The USBL measures the DoA, the Doppler speed, and the slant range, but to test our approach, we only utilize the DoA and Doppler speed measurements. The parameters of the instruments are shown in TABLE \ref{experiment}.}

\begin{figure}[htbp]
	\centering
	\subfloat[$\delta_y$ = \SI{0}{\degree} ]{\includegraphics[width=0.49\linewidth]{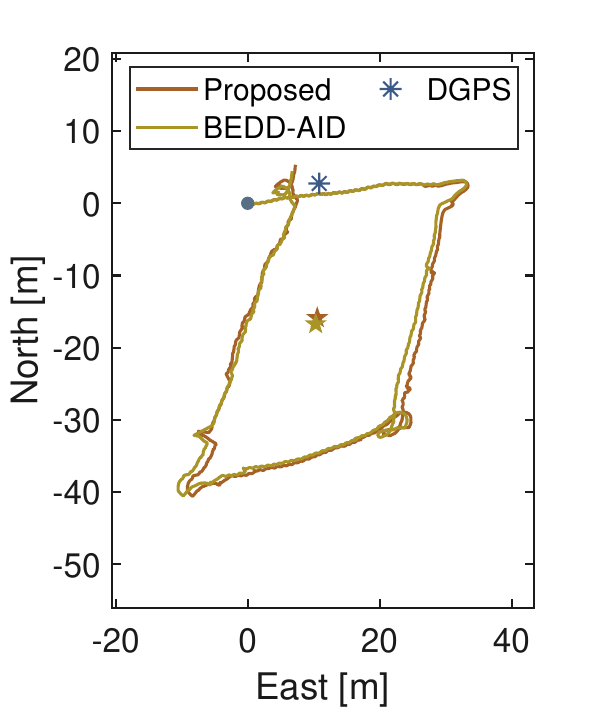}}
	\subfloat[$\delta_y$ = \SI{-9}{\degree}]{\includegraphics[width=0.49\linewidth]{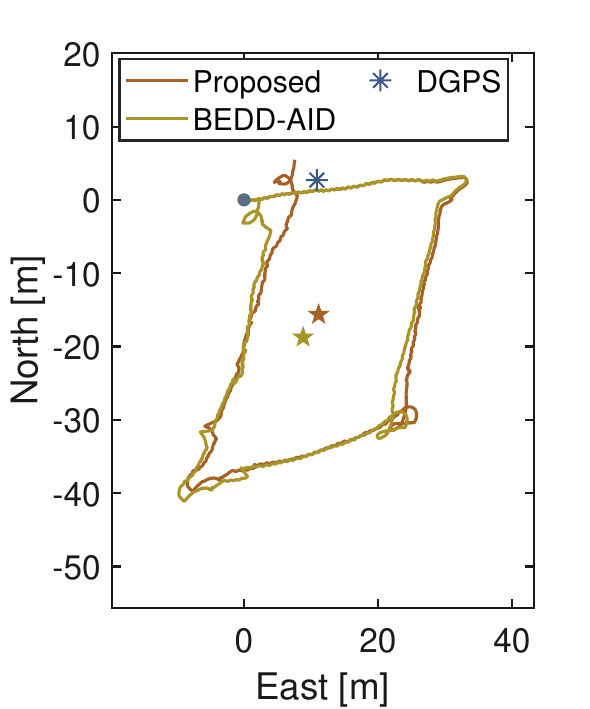}}
	\quad
	\caption{The navigation solutions with respect to different misalignment angles in the field tests. Marker \textcolor{red1}{$\star$} denotes the beacon position estimated by the proposed method, while marker \textcolor{yellow1}{$\star$} indicates the result of BEDD-AID. The origins of the trajectories are aligned (denoted by \textcolor{black1}{$\bullet$}). Marker \textcolor{blue1}{$\ast$} indicates the DGPS fix of the last position. $\delta_y$ is the added yaw misalignment angle. }
	\label{figFieldTraj}
\end{figure}

\begin{figure}[htbp]
	\centering
	\subfloat[Final error of navigation ]{\includegraphics[width=0.49\linewidth]{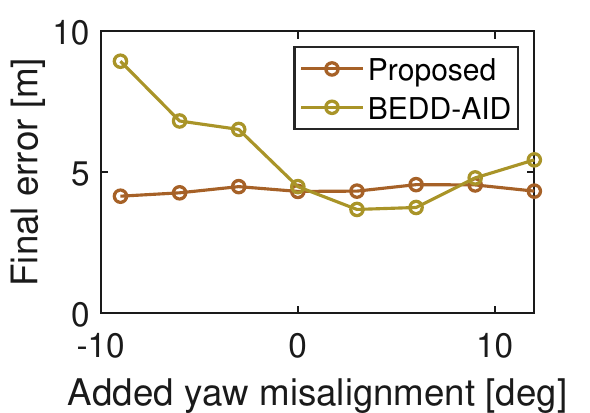}}
	\subfloat[Alignment results versus added bearing offsets]{\includegraphics[width=0.49\linewidth]{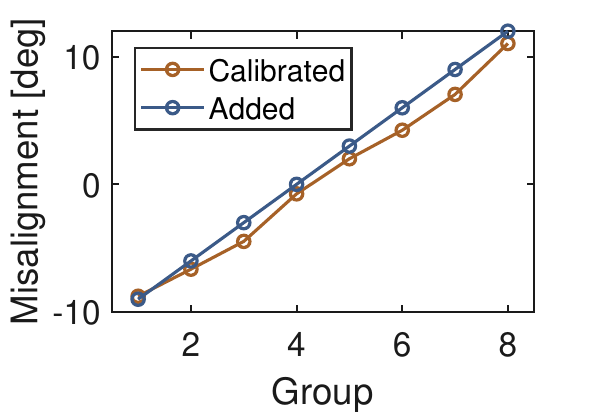}}
	\quad
	\caption{Navigation and alignment results for different added misalignment angles in the field tests. The final navigation error is calculated by taking the DGPS fix of the last position as ground truth.}
	\label{figFieldAlign}
\end{figure}

\begin{table}[htbp]
	\renewcommand\arraystretch{1.2}
	\setlength{\belowcaptionskip}{0pt}%
	\caption{EXPERIMETNTAL SETUP}
	\begin{center}
		\begin{tabular}{p{0.19\linewidth}p{0.5\linewidth}c}
			\toprule[1.5pt]
			Instrument&Parameter & Value\\
			\midrule
			\multirow{2}{*}{AHRS (\SI{20.0}{\hertz})}&Roll/Pitch measurement noise [$^{\circ}$] & 0.2\\ 
			
			&Yaw measurement noise [$^{\circ}$]&1.0\\
			\midrule        
			\multirow{2}{*}{DVL (\SI{2.0}{\hertz})}&Acoustic frequency [kHz]& 300\\
			&Velocity measurement noise [m/s]&0.02 \\
			\midrule        
			\multirow{3}{*}{USBL (\SI{0.1}{\hertz})}&Frequency band [kHz]&18-34\\
			&Bearing resolution [$^{\circ}$] & 0.1 \\           
			&Slant range accuracy [m]&0.01\\
			\bottomrule[1.5pt]
			
		\end{tabular}
		\begin{tablenotes}
			\footnotesize
			\item The Doppler speed accuracy is not specified, roughly better than 0.1 m/s.
		\end{tablenotes}
	\end{center}
	\label{experiment}
\end{table} 
		\addtolength{\textheight}{-0.0cm}   
		{}{
 To evaluate the tolerance to the sensor misalignment, we set the yaw misalignment by manually biasing the DoA measurements. Specifically, we add different angles to the bearing measurements to set different misalignments between the body frame and the acoustic array frame. The results are presented in Fig. \ref{figFieldTraj} and Fig. \ref{figFieldAlign}.
When the added sensor misalignment is minimal, 
the final error of BEDD-AID is very close to the proposed method. However, our method is much more robust to significant sensor misalignments since its navigation error is almost the same under different misalignment angles. Note that the ground truth of the alignment error is unavailable since we do not know the original misalignment angle before manually adding an offset. Nevertheless, it can be seen from Fig. \ref{figFieldAlign} that the yaw misalignment angles estimated by our method have the same changing trends as the manually added misalignment angles. Therefore, our method is capable of estimating the misalignment angles online.}

\section{Conclusion}
\label{sectConclusion}
{}{
	This paper presents a novel approach to online beacon localization and sensor alignment for low-power AUV navigation. The experimental results indicate that the proposed method is tolerant to sensor misalignment and is robust to shallow elevation angles. }
	{}{
	According to some recent literature on frequency difference of arrival (FDOA)-based passive source localization \cite{cameron2018fdoa}, it is possible to position the acoustic beacon with only Doppler measurements. In future work, we will explore the possibility of utilizing only Doppler measurements for beacon localization before sensor alignment, for example, by improving the AUV speed and planning the trajectory. This way, beacon localization can be decoupled from sensor alignment, thus enhancing the system's observability.}
	{}{
	Additionally, we will conduct more real-world experiments to evaluate the reliability of the proposed approach and make a comprehensive comparison with other methods. Also, its performance in cooperative navigation will be verified.  }

		\bibliographystyle{IEEEtran}
		\bibliography{IEEEabrv,IEEEexample}

	\end{document}